\documentclass[11pt]{article}

\usepackage[top=1in, right=1in, left=1in, bottom=1in]{geometry}

\usepackage[in]{fullpage}  

\usepackage[utf8]{inputenc} 
\usepackage[T1]{fontenc}    
\usepackage[colorlinks=true, linkcolor=blue, citecolor=blue]{hyperref} 
\usepackage{url}            
\usepackage{booktabs}       
\usepackage{amsfonts}       
\usepackage{nicefrac}       
\usepackage{microtype}      

\usepackage{amsmath}
\usepackage{amsthm}
\usepackage{amssymb}
\usepackage{graphicx}
\usepackage{algorithm,float}
\usepackage{appendix}
\usepackage{algorithmic}
\usepackage{color}
\usepackage{enumitem}
\usepackage{psfrag}
\usepackage{bm}
\usepackage{subfigure}

\newtheorem{condition}{Condition}
\newtheorem{defn}{Definition}
\newtheorem{lem}{Lemma}
\newtheorem{thm}{Theorem}

\newtheorem{rem*}{Remark}
\global\long\def\E{\mathbb{E}}%
\global\long\def\mP{\mathcal{P}}%
\global\long\def\mR{\mathcal{R}}%
\global\long\def\R{\mathbb{R}}%
\global\long\def\Z{\mathbb{Z}}%

\global\long\def\mA{\mathcal{A}}
\global\long\def\mS{\mathcal{S}}

\global\long\def\mF{\mathcal{F}}

\global\long\def\P{\Pr}
\global\long\def\E{\mathbb{E}}
\global\long\def\P{\mathbb{P}}
\global\long\def\hV{\widehat{V}}
\global\long\def\hQ{\widehat{Q}}
\global\long\def\err{{\sf err}}

\global\long\def\lip{\zeta}

\global\long\def\Sn{S_{\sf next}}

\title{Stable Reinforcement Learning with Unbounded State Space}

\author{%
  Devavrat Shah \\
  EECS, MIT \\
  \texttt{devavrat@mit.edu} \\
  \and
  Qiaomin Xie \\
  IEOR, Cornell University \\
  \texttt{qiaomin.xie@cornell.edu} \\
  \and
  Zhi Xu \\
  EECS, MIT \\
  \texttt{zhixu@mit.edu} \\
}

\begin{document}

\date{}
\maketitle

\begin{abstract}

We consider the problem of reinforcement learning (RL) with unbounded state space motivated by the classical problem of scheduling in a queueing network. Traditional policies as well as error metric that are designed for finite, bounded or compact state space, require infinite samples for providing any meaningful performance guarantee (e.g. $\ell_\infty$ error) for unbounded state space. That is, we need a new notion of performance metric. As the main contribution of this work, inspired 
by the literature in queuing systems and control theory, we propose {\em stability} as the notion of ``goodness'': 
the state dynamics under the policy should remain in a {\em bounded} region with high probability. As a proof of concept, we propose an RL policy using Sparse-Sampling-based Monte Carlo Oracle and argue that it satisfies the stability property as long as the system dynamics under the optimal policy respects a Lyapunov function. The assumption of existence of a Lyapunov function is not restrictive as it is equivalent to the positive recurrence or stability property of any Markov chain, i.e., if there is any policy that can stabilize the system then it must posses a Lyapunov function. And, our policy does not utilize the knowledge of the specific Lyapunov function. To make our method sample efficient, we provide an improved, sample efficient 
Sparse-Sampling-based Monte Carlo Oracle with Lipschitz value function that may be of interest in its own right. 
Furthermore, we design an adaptive version of the algorithm, based on carefully constructed statistical tests, which finds the correct tuning parameter automatically.

\end{abstract}

\section{Introduction}

We consider the problem of reinforcement learning (RL) for controlling an unknown dynamical systems with an unbounded state space. Such problems are ubiquitous in various application domains, as exemplified by scheduling for networked systems. As a paradigm for learning to control dynamical systems, RL has a rich literature. In particular, algorithms for the setting with finite, bounded or compact state spaces has been well studied, with both classical asymptotic results and recent non-asymptotic performance guarantees. However, literature on problems with unbounded state space is scarce, with few exceptions such as linear quadratic regulator~\cite{abbasi2011regret,dean2019safely}, where the structure of the dynamics is {\em known}. 
Indeed, the unboundedness of the state space presents with new challenges for algorithm or policy design, 
as well as analysis of policy in terms of quantifying the ``goodness''. 

\begin{figure}[ht]
    \centering
    \includegraphics[width=0.35\textwidth]{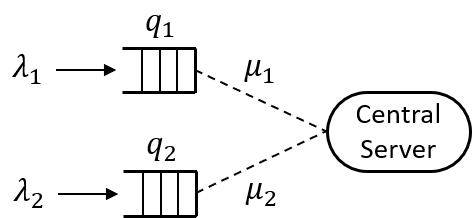}
    \vspace{-0.04in}
    \caption{A two queue network: arrival, service rates for queue $i \in \{1,2\}$ are $\lambda_i$, $\mu_i$, respectively.}
    \label{fig:queue_example}
\end{figure}

\medskip
\noindent{\bf A motivating example.} To exemplify the challenges involved, we consider a simple example of discrete-time queueing system with two queues as shows in Figure~\ref{fig:queue_example}: Jobs arrives to queue $i \in \{1, 2\}$ per Bernoulli process with rate $\lambda_i \in (0,1)$. A central server can choose to serve job from one of the queues at each time, and if a job from queue $i$ is chosen to serve, it departs the system with probability $\mu_i \in (0,1)$. That is, in effect 
$\rho_i = \lambda_i / \mu_i$ amount of ``work'' arrives to queue $i$ while total amount of work the system can do is 
$1$. The state of the system is $q = (q_1, q_2)$ with $q_i$ representing number of jobs in the $i$th queue. 

The evolution of the system is controlled by a scheduling decision that specifies which queue $i\in \{1,2\}$ to serve at each time. Viewed as a Markov decision process (MDP), the state space is $\mS = \{0,1,\ldots\}\times\{0,1,\ldots\}  $ and the action space is $ \mA = \{1,2\}$. Operating under a Markov policy $\pi: \mS \to \mA$, the server will serve queue $i=\pi(q)$ at state $q$. The problem of stochastic control is to identify a policy that optimizes a given criterion (e.g., average or discounted total queue lengths). Finding such a policy using empirical data is the question of Reinforcement Learning (RL). 

\noindent{\bf Solutions that may not work.} In traditional RL approach, the policy is trained {\em offline} using finitely many samples for finite, bounded or compact state spaces and then it is deployed in 
{\em wild} without further changes. A natural adaption of such an approach is by restricting the RL policy to a finite subset of the state space chosen appropriately or arbitrarily. Examples of such algorithms include model-based methods (such as UCRL/PSRL~\cite{jaksch2010UCRL,osband2013PSRL}) that estimate the transition probabilities of the system and then solve the dynamic programming problem on the estimated system, and  model-free methods (such as TD/Q-Learning~\cite{sutton1998reinforcement}) that directly estimate the value function. 

However, even in our simple, motivating example, the system will reach a state $q$ not contained in the training data with non-zero probability. The estimate for $q$'s transition probabilities and value function will remain at their initial/default values (say 0). With such an uninformative estimate, the corresponding policy will be independent of the state $q$. And it is likely that the policy may end up serving empty queue with a nonzero probability. This might cause the queues to grow unboundedly with strictly positive probability. Clearly, more sophisticated approaches to truncate system are not going to help as they will suffer from similar issue. 

An alternative to truncation is to find ``lower dimensional structure'' through functional approximation, 
e.g., by parametrizing the policy $\pi$ within some function class (such as linear functions or neural networks). For this approach to work, the function class must be expressive enough to contain a stable policy. However, it is not at all clear, \emph{a priori}, which parametric function class has this property, even for the simple 
example in Figure~\ref{fig:queue_example}. This challenge is only exacerbated in more complicated systems. Although some approximation architectures work well empirically~\cite{mao2018variance,mao2019learning}, there is no rigorous performance guarantee in general.

\medskip
\noindent{\bf Challenges.} To sum up, the traditional RL approaches  for finite, bounded or compact  state space 
are not well suited for systems with unbounded state space. Approaches that rely on {\em offline}
training only are bound to fail as system will reach a state that is not observed in finitely many samples during offline training and hence, there is no meaningful guidance from the policy. 
Therefore, to learn a reasonable policy with an unbounded state space, the policy ought to be updated whenever a new scenario is encountered. That is, 
unlike traditional RL, we need to consider {\em online} policies,  i.e., one that is continually 
updated upon incurring new  scenarios, very much like that in the context of Multi-arm bandit (cf.~\cite{bubeck2012regret}).

Another challenge is in analyzing or quantifying ``goodness'' of such a policy. Traditionally, the ``goodness'' of 
an RL policy is measured in terms of the error induced in approximating, for example, the optimal value function over the entire state space; usually measured through $\|\cdot\|_\infty$ norm error bound, cf.~\cite{Kearns1999finite,bertsekas2017dynamic}. Since the state space is unbounded, expecting a good approximation of the optimal value function over the entire state space is not a meaningful measure. Therefore, we need an alternative to quantify the ``goodness'' of a policy. 

\medskip
\noindent{\bf Questions of interest. } In this work, we are  interested in the following questions: 
(a) What is the appropriate ``goodness'' of performance for a RL policy for unbounded state space?
(b) Is there an online, data-driven RL policy that achieves such ``goodness''? and if so, (c) How does
the number of samples required per time-step scale? 

\medskip
\noindent{\bf Our contributions.} 
Motivated by the above considerations, we consider 
discounted Markov Decision Processes with an unbounded state space and a finite action space, under a generative model. 

\medskip
{\em New Notion of Stability.}
As the main contribution, we introduce notion of {\em stability} to quantify ``goodness'' of RL policy for
unbounded state space inspired by literature in queueing systems and control theory. Informally, an RL
policy is stable if the system dynamics under the policy returns to a finite, bounded or compact subset of
the system infinitely often --- in the context of our example, it would imply that queue-sizes remain finite
with probability $1$. 

\medskip
{\em Stable RL Policy.} As a proof of concept, we present a simple RL policy using a Sparse Sampling Monte Carlo Oracle
that is stable for any MDP, as long as the optimal policy respects a Lyapunov function with {\em drift} 
condition (cf.~Condition~\ref{cond:lyapunov}). Our policy {\em does not} require knowledge of or access to such a Lyapunov function. It recommends 
an action at each time with finitely many {\em simulations} of the MDP through the oracle. That is, the policy is {\em online} and guarantees stability for each trajectory starting {\em without}  
any prior training.
The number of samples 
required at each time step scales as 
$O\Big(\big(\frac{1}{\alpha^4} \log^2 \frac{1}{\alpha}\big)^{O(\log \frac{1}{\alpha})}\Big),$ 
where $-\alpha <0$ is the drift in Lyapunov function. To  our best knowledge, ours is the first online 
RL policy that is {\em stable} for generic MDP with unbounded state space. 

\medskip
{\em Sample Efficient Stable RL Policy.} To further improve the efficiency, for MDPs with Lipschitz optimal value function, we propose a modified Sparse Sampling Monte Carlo Oracle for which the number of samples required at each time step scales as 
 $O\Big(\frac{1}{\alpha^{2d+4}} \log^{d+1} \frac{1}{\alpha}\Big),$ where $d$ is the dimension
of the state space. That is, the sample complexity becomes polynomial in $1/\alpha$ 
from being super-polynomial with the vanilla oracle. The efficient oracle utilizes the minimal structure of smoothness in the optimal value function and should be of interest in its own right as it provides
improvement for {\em all} policies in the literature where such an oracle plays a key role, e.g., \cite{kearns2000fast,parkes2005mechanism}.

\medskip
{\em Adaptive Algorithm Based on Statistical Tests.}
While the algorithm does not require knowing the Lyapunov function itself, it does have a parameter whose optimal value depends on the drift parameter of the Lyapunov function. Therefore, we further develop an adaptive, agnostic version of our algorithm that automatically searches for an appropriate tuning parameter. We establish that either this algorithm discovers the right value and hence ensures stability, or the system is near-stable in the sense that $||s_t||/\log^2t=O(1)$ as $t\rightarrow\infty$. The near-stability is a form of sub-linear regret. For example, in the context of a queueing system, 
this would correspond to queues growing as $O(\log^2 t)$ with time in contrast to $O(1)$ queues for stable (or optimal) policy. 
Further, in the context of queueing system, it would imply ``fluid'' or ``rate'' stability cf.~\cite{dai1995stability,dai1995unified} --- to the 
best of our knowledge, {this is first such general RL policy for generic queueing systems with such a property.}

\begin{table}[t]
\setlength{\tabcolsep}{5.5pt}
\caption{Comparison with prior work on safety and stability. }
\vspace{-0.25in}
\label{tab:one}
\begin{center}
\resizebox{\textwidth}{!}{
\begin{tabular}{c|c|c}
\toprule[1.3pt]
  & Settings/Conditions & Guarantees\\
 \midrule
 { Ours}   & { Unbounded space; Unknown dynamics; Existence of unknown Lyapunov function} &    { Stochastic Stability}\\
  \hline
 \cite{dean2019safely}   & Linear dynamics; Unknown parameters; Quadratic cost (LQR) &    Constraints on state \& action\\
 \hline
  \cite{turchetta2016safe} & Finite space; Deterministic, known dynamics;  Gaussian safety function & Constraints on state \& action \\
  \hline
  \cite{yu2019convergent}   & Unknown dynamics; Compact parametrized policy class &  Expected constraint costs (CMDP)  \\
  \hline
  \cite{vinogradska2016stability} & Gaussian process dynamics; Unknown parameters  & Control-theoretic stability\\
  \hline
  \cite{berkenkamp2017safe} & Compact space; Deterministic, partially known dynamics; Access to Lyapunov func. & Control-theoretic stability\\

\bottomrule[1.3pt]
\end{tabular}}
 \vspace{-0.22in}
\end{center}
\end{table}

\medskip
\noindent{\bf Related work.} The concept of stability introduced in this paper is related to the notion 
of \emph{safety} in RL, but with crucial differences. Various definitions of safety exist in literature~\cite{pecka2014safe,garcia2015comprehensive}, such as 
hard constraints on individual states and/or actions~\cite{garcia2012safe,hans2008safe,dalal2018safe,dean2019safely,koller2018learning}, 
expected cumulative constraint costs formulated as Constrained MDP~\cite{achiam2017constrained,chow2018lyapunov,yu2019convergent} and  control-theoretic notions of stability~\cite{berkenkamp2016safe,vinogradska2016stability,berkenkamp2017safe}.
Importantly, in our work, stability is defined in terms of the positive recurrence of Markov chains, which cannot be immediately written as constraints/cost over the states and actions. In particular, our stability notion captures long-term behaviors of the system---it should eventually stay in a desirable, bounded region  with high probability. In general, there does not exist an action that immediately drives the system back to that region; learning a policy that achieves so in the long run is non-trivial and is precisely our goal. Overall, we believe this new notion of stability provides a generic, formal framework for studying  RL with unbounded state space. 
In Table \ref{tab:one}, we provide a concise comparison with some prior work, and we refer the readers to Appendix~\ref{app:related} for detailed discussions. 

\medskip
\noindent{\bf Organization.} Section~\ref{sec:formulation} introduces the framework and formally defines the notion of stability. In Section~\ref{sec:online_RL}, we provide three online algorithms and establish their stability guarantee. The proofs of our main theorems are provided in Appendices~\ref{sec:analysis}--\ref{sec:analysis3}.

\section{Setup and Problem Statement}
\label{sec:formulation}

\subsection{Markov Decision Process and Online Policy}
\noindent{\bf Markov Decision Process (MDP).} We consider a discrete-time discounted Markov decision
process (MDP) defined by the tuple $(\mS,\mA,\mP,\mR,\gamma),$ where $\mS$ and $\mA$ are the 
state space and action space, respectively,  $\mP\equiv p(s'|s,a)$ is the Markovian transition kernel,
$\mR=\mR(s,a)$ is the reward function, and $\gamma\in(0,1)$ is a discounted factor. At time $t \in \{0,1,\dots\}$, 
the system is in state $s_t \in \mS$;  upon taking action $a_t\in\mA$, the system transits to $s_{t+1} \in \mS$ 
with probability $p(s_{t+1} | s_t,a_t)$  and generates a reward $\mR(s_t,a_t).$  At time $t$, the action $a_t$ is
chosen as per some policy $\pi^t \equiv [\pi^t(a | s), ~ a\in \mA, s \in \mS],$ where $\pi^t(a | s)$ represents the
probability of taking action $a_t = a$ given state $s_t = s$. If $\pi^t = \pi$ for all $t$, then it is called a stationary policy. For 
each stationary policy, we define the standard value function and Q-function, respectively, as 
$V^{\pi}(s)  = \E_{\pi}\big[\sum_{t=0}^{\infty}\gamma^{t}\mR(s_{t},a_{t})|s_{0}=s\big], \forall s \in \mS, $ and 
$Q^{\pi}(s, a)  =\E\big[ \mR(s_0, a_0) + \gamma V^{\pi}(s_1) | s_0=s, a_0=a \big], \forall s \in \mS, \forall a\in \mA$.
An optimal stationary policy $\pi^{*}$ is the policy that achieves optimal value, i,e., 
$V^{\pi^{*}}(s) =\sup_{\pi}V^{\pi}(s),\forall s\in\mS$. 
Correspondingly, 
$Q^{\pi^{*}}(s,a)  = \E\big[ \mR(s_0, a_0) + \gamma V^{\pi^{*}}(s_1) \big | s_0=s, a_0=a\big], \forall s \in \mS, \forall a\in \mA$.
It is well understood that such an optimal policy and associated value/Q functions, $V^* \equiv V^{\pi^{*}}$ and $Q^* \equiv Q^{\pi^{*}}$, 
exist in a reasonable generality; cf.~\cite{bertsekas2017dynamic}. We focus on MDPs satisfying the following condition.
\begin{condition}\label{cond:1}
Action space $\mA$ is finite, and state space $\mS \subseteq \R^{d}$ is unbounded with some $d \geq 1$.
\end{condition}
We assume that the reward function $\mathcal{R}$ is bounded and takes value in $[0,R_{\max}]$. Consequently, for {\em any} 
policy, $V$ and $Q$ function are bounded and take value in $[0,V_{\max}],$ 
where $V_{\max} \triangleq R_{\max} /(1 - \gamma)$. Breaking ties randomly, we shall restrict to optimal policy $\pi^*$ of the following form: 
for each $s \in \mS$, 
\begin{align*}
\pi^{*}(a | s)  & = \begin{cases}
			\frac{1}{|\mA^{*}(s)|} & \mbox{~if~}  a\in\mA^{*}(s), \\
			0 & \mbox{otherwise,}
		          \end{cases}
\end{align*}
where $\mA^{*}(s)=\arg\max_{a\in\mA}Q^{*}(s,a)$. Define 
$\Delta_{\min}(s)=\max_{a\in\mA}Q^{*}(s,a)-\max_{a\in\mA\setminus\mA^{*}(s)}$ $Q^{*}(s,a).$
For a given state $s \in \mS$, if not all actions are optimal, then $\Delta_{\min}(s) > 0$. Denote the minimum gap by $\Delta_{\min} \triangleq \inf_{s\in\mS}\Delta_{\min}(s)$. Throughout the paper, we assume that $\Delta_{\min} > 0 $.

\medskip
\noindent{\bf System dynamics and online policy.} In this work, our interest is in designing online policy starting with no prior information. Precisely, the system starts with arbitrary initial state $s_0 \in \mS$ and initial policy $\pi^0$. 
At time $t \geq 0$, given state $s_t$, action $a_t \in \mA$ is chosen with probability $\pi^t(a_t | s_t)$ leading to 
state $s_{t+1} \in \mS$ with probability $p(s_{t+1} | s_t, a_t)$. At each time $t$, policy $\pi^t$ is decided using 
potentially a finite number of simulation steps from the underlying MDP, in addition to the historical information observed till time $t$. In this sense, the policy is {\em online}.

\subsection{Stability}
We desire our online policy to have a {\em stability} property, formally defined as follows.

\begin{defn}[Stability] 
\label{def:stability}	
We call policy $\{\pi^t, ~t \geq 0\}$ stable if for any $\theta \in (0, 1)$, 
there exists a bounded set $\mS(\theta)\subset \mS$ such that the following are satisfied:
\begin{enumerate}
    \setlength{\itemsep}{0pt}
    \setlength{\parskip}{2pt}
    \item Boundedness: 
    \begin{equation}
        \limsup_{t\rightarrow\infty}\P\big(s_t\notin \mS(\theta)\mid s_0 = s\big)\leq \theta,\quad\forall\:s\in\mS.
    \end{equation}
    \item Recurrence: Let $T(s,t, \theta)=\inf\{k\geq 0: s_{t+k}\in \mS(\theta) \mid s_t=s\}$.\footnote{$\inf$ of empty
    set is defined as $\infty$.} Then
    \begin{equation}
       \sup_{t\geq 0}\E\big[T(s,t,\theta)|s_t = s\big] < \infty,\quad\forall\: s\in \mS.
    \end{equation}
\end{enumerate}
\end{defn}

In words, we desire that starting with {\em no prior information}, the policy learns as it goes and manages to
retain the state in a finite, bounded set with high probability. It is worth remarking that the above stability property is similar to the recurrence property for Markov chains. 

\medskip
\noindent{\bf Problem statement.} Design an {\em online} stable policy for a given, discounted cost MDP. 

\medskip
\noindent{\bf MDPs respecting Lyapunov function.} As we search for a stable policy for MDP, if the optimal stationary policy for MDP is stable, then the resulting system dynamics under optimal policy is a time homogeneous Markov chain that is positive recurrent. The property of positive recurrence is known to be equivalent to the existence of the so-called {\em Lyapunov} function; see \cite{mertens1978necessary}. In particular, if there exists a policy for the MDP under which the resulting Markov chain is positive recurrent, then this policy has a Lyapunov function. These observations motivate us to restrict attention to MDPs with the following property. 
\begin{condition}\label{cond:lyapunov} 
The Markov chain over $\mS$ induced by MDP operating under the optimal policy $\pi^*$ respects a
Lyapunov function $L:\mS\rightarrow\R_{+}$ such that $\liminf_{\Vert s\Vert_2 \to \infty} L(s) = \infty$, 
$\sup\{\|s\|_2 : L(s) \leq \ell \} < \infty$ for all finite $\ell$, and such that for some $\nu, \nu', B, \alpha > 0$, for any $t \geq 0$: 
\begin{enumerate}
 \setlength{\itemsep}{0pt}
 \setlength{\parskip}{4pt}
\item (Bounded Increment) For every $s_{t}\in\mS$ and every $a\in\mA$,
\begin{align}\label{eq:bounded.increment}
\left|L(s_{t+1})-L(s_{t})\right|\leq\nu, \quad \|s_{t+1} - s_t\| \leq \nu',
\end{align}
for all possible next states $s_{t+1}$ with $ p(s_{t+1}|s_{t},a) >0$.\footnote{$p(\cdot|s_t,a)$ should be understood as the density of the conditional distribution of the next state with respect to the 
Lebesgue measure (for continuous state space) or the counting measure (for discrete state space).}
\item (Drift Condition) For every $s\in\mS$ such that $L(s)>B,$
\begin{align}\label{eq:drift}
\E\big[L(s_{t+1})-L(s_{t})|s_{t}=s\big]\leq-\alpha.
\end{align}
\end{enumerate}
\end{condition}

Recall our motivating example of a queueing system with two queues. It is well known and 
can be easily verified that for the serve-the-longest-queue policy, $L(q_1, q_2) = \sqrt{q_1^2 + q_2^2}$ 
serves as a Lyapunov function satisfying Condition \ref{cond:lyapunov} as long as $\rho_1 + \rho_2  < 1$. 
As explained above,
the requirement of MDP respecting Lyapunov function under optimal policy is not restrictive. 
We further note that our policy {\em will not} require {\em precise} knowledge of or access to such a Lyapunov function.

\section{Online Stable Policy}
\label{sec:online_RL}

We present our main results in this section. First, we describe a stationary, online policy that is stable 
under Conditions~\ref{cond:1} and~\ref{cond:lyapunov}. This policy is simple and provides key intuition behind our approach, though being sample inefficient. Next, we present an efficient version thereof that utilizes the minimal structure of Lipschitz optimal value function (cf.~Condition~\ref{cond:lip.val}). Finally, we design an adaptive version of our algorithm that automatically discover the appropriate tuning parameter without knowing the value of the drift parameter of the Lyapunov function.

\subsection{Sample Inefficient Stable Policy}\label{ssec:vanilla.policy}

\noindent{\bf Overview of policy.} 
At each time $t \geq 0$, given the state $s_t \in \mS$, an action $a_t \in \mA$ is chosen by sampling from a distribution over $\mA$. The distribution is determined by an online planning algorithm that uses finitely many simulations of MDP performed at time $t$ and depends only on the input state $s_t$. 
Therefore, the policy is stationary. Precisely, using Monte Carlo Oracle with Sparse Sampling for parameters $C, H > 0$ (details below, adapted from \cite{kearns2002sparse}), we produce $\hat{Q}(s_t, a)$ as an estimation of $Q^*(s_t, a)$, for each action $a \in \mA$. We use Boltzman distribution with
parameter $\tau > 0$ for sampling action $a_t$: 
   \begin{align*}
       \hat{\pi}(a | s_t) & = \frac{e^{\hat{Q}(s_t, a)/\tau}}{\sum_{a'}e^{\hat{Q}(s_t, a')/\tau}},\:\forall\: a\in\mathcal{A} .
   \end{align*}

\medskip
\noindent{\bf Sparse Sampling Monte Carlo Oracle.} We describe the Monte Carlo Oracle based on sparse sampling~\cite{kearns2002sparse}. Given an input state $s \in \mS$ with two integer valued parameters, $C, H > 0$,
and an estimate of value function $\hV$, it produces an estimate of $Q^*(s, a)$ for all $a \in \mA$. 
The error in estimate depends on $C, H$ and error in value function $\hV$. With larger
values of $C$ and $H$, the estimation error decreases but the number of simulations of MDP required increases. 
Next we provide details of the algorithm. 

Sparse sampling oracle constructs a tree of depth $H$ representing a partial $H$-step look-ahead of MDP starting with the input state $s_0 = s$ as its root. The tree is constructed iteratively as follows: at any state node $s'$ in the tree (initially, root node $s_0$), for each action $a \in \mA$, sample $C$ times 
the next state of MDP for the state-action pair $(s', a)$.
Each of the resulting next states, $C$ for each $a \in \mA$, are placed as children nodes, 
with the edge between $s'$ and the child node labeled by the generated reward. 
The process is repeated for all nodes of each level until reaching a depth of $H$. 
That is, it builds an $|\mA|C$-array tree of depth $H$ representing a partial $H$-step look-ahead 
tree starting from the queried state $s_0$, and hence the term \emph{sparse sampling} oracle.

To obtain an estimate for the $Q$-values associated with $(s_0, a), ~a \in \mA$, we start by assigning value $0$
to each leaf node $s_{\sf leaf}$ at depth $H$, i.e., $\hV^H(s_{\sf leaf}) = 0$.
These values, together with the associated rewards on the edges, are then backed-up to find 
estimates of $Q$-values for their parents, i.e., nodes at depth $H-1$.  The estimate for the $Q$-value 
of a parent node $s$ and an action $a$ is just a simple average over $a$'s children, i.e., 
$\hQ^{H-1}(s,a)=R_{sa}+\gamma \frac{1}{C}\sum_{s'\in \Sn(s, a, C)}\hV^H(s'),$ where $R_{sa}$ 
is the reward on the edge of action $a$, and $\Sn(s, a, C)$ is the set of $a$'s children nodes. 
The estimate for the value of state $s$ is given by $\hV^{H-1}(s)=\max_a \hQ^{H-1}(s,a).$
The process is recursively applied from the leaves up to the root level to find estimates of 
$\hat{Q}^0(s_0,a)$ for the root node $s_0$. 
\begin{lem}[Oracle Guarantee]
\label{lem:sparse}
Given input state $s \in \mS$ and $\delta > 0$, under the Sparse Sampling Monte Carlo  Oracle, we have
\begin{align}
    \max_{a \in \mA} |Q^*(s,a) - \hQ^0(s,a)| &\leq 2V_{\max}\delta, 
\end{align}
with probability at least $1-\delta$, with choice of parameters  
\begin{align}
    H &= \lceil \log_\gamma (\delta) \rceil, 
\qquad     C = \frac{2\gamma^2}{\delta^2(1-\gamma)^2}\Big(2H\log\frac{2\gamma^2|\mA|H }{\delta^2(1-\gamma)^2} + \log\frac{2}{\delta}\Big). \label{eq:spare_oracle_C}
\end{align}

The number of simulation steps of MDP utilized are $O\big((|\mA|C)^H\big)$, which as a function of $\delta$ scales
as $O\Big(\big(\frac{1}{\delta^2}\log^2 \frac{1}{\delta}\big)^{O(\log \frac{1}{\delta})}\Big)$.
\end{lem}
We shall omit the proof of Lemma~\ref{lem:sparse} 
as it follows directly from result in~\cite{kearns2002sparse}. Further, 
we provide proof of Lemma~\ref{lem:sparse.mod} which establishes similar guarantees for a modified sample-efficient 
Sparse Monte Carlo Oracle. 

\subsubsection{Stability Guarantees}
\noindent{\bf Stability.} We state the result establishing stability of the stationary policy described above under appropriate choice of parameters $\delta, \tau, C$ and $H > 0$. The proof is provided in Appendix~\ref{sec:analysis}.

\begin{thm}
\label{thm:main.1}
Suppose that the MDP of interest satisfy Conditions \ref{cond:1} and \ref{cond:lyapunov}. Then, with 
$\delta = \tau^2$, $\tau$ as 
\begin{align}\label{cond:tau}
\tau & \leq \tau(\alpha) \triangleq \min\Bigg\{ \sqrt{\frac{\alpha}{24\nu}}, \frac{1-\gamma}{8R_{\max}},
\frac{(1-\gamma)\alpha}{48 R_{\max}|\mA|^2 \nu},  
 \frac{\Delta_{\min}}{\ln \Big(\frac{24\nu |\mA|}{\alpha}\Big)}\Bigg\},
\end{align}
and $C, H > 0$ chosen as per Lemma \ref{lem:sparse} for a given $\delta$, the resulting stationary policy is stable. 
\end{thm}

\subsection{Sample Efficient Stable Policy}\label{ssec:efficient.policy}

\noindent{\bf The Purpose.} Theorem \ref{thm:main.1} suggests that, as a function of $\alpha$, $\delta (=\tau^2)$ 
scales as $\min(\alpha, \alpha^2)$. Indeed, as $\alpha$ becomes larger, i.e., the negative drift for Lyapunov function 
under optimal policy increases, system starts living in a smaller region with a higher likelihood. The challenging regime 
is when $\alpha$ is small, formally $\alpha \to 0$. Back to our queueing example, this corresponds to what is 
known as {\em Heavy Traffic} regime in queueing systems. Analyzing complex queueing systems in this
regime allows understanding of fundamental ``performance bottlenecks'' within the system and subsequently 
understand the properties of desired optimal policy. Indeed, a great deal of progress has been made over more
than past four decades now; see for example \cite{HT1, HT2, HT3}. 
Back to the setting of MDP with $\alpha \to 0$: Based on Theorem \ref{thm:main.1}, the number of samples required per time step for the policy 
described in Section \ref{ssec:vanilla.policy} scales as 
$O\Big(\big(\frac{1}{\alpha^4} \log^2 \frac{1}{\alpha}\big)^{O(\log \frac{1}{\alpha})}\Big)$. That is, the policy
is {\em super-polynomial} in $1/\alpha$. 

\medskip
\noindent{\bf Minimal Structural Assumption.} In what follows, we describe a stable policy with the number of samples
required scaling {\em polynomially} in $1/\alpha$, precisely $\tilde{O}(1/\alpha^{2d+4})$ where $d$ is the dimension of the 
state-space $\mS$. This efficiency is achieved due to minimal structure in the optimal value 
function in terms of Condition \ref{cond:lip.val}, which effectively states that the value function is Lipschitz. Specifically,  
we provide an efficient Sparse Sampling Monte Carlo Oracle exploiting the 
Bounded Increment property \eqref{eq:bounded.increment} in Condition \ref{cond:lyapunov} along with 
Lipschitzness of optimal value function as described in Condition \ref{cond:lip.val}. 
We remark that for learning with continuous state/action space, smoothness assumption such as the Lipschitz continuity is natural and typical~\cite{antos2008fitted,shah2019nonasymptotic,shah2018qlnn,dufour2012approximation}. 

\begin{condition}\label{cond:lip.val} 
Let $\mS = \mathbb{R}^d$. The optimal value function $V^*: \mS \to \R$ is a Lipschitz continuous function. Precisely, there exists
constant $\lip > 0$ such that for any $s, s' \in \mS$, 
\begin{align}
|V^*(s) - V^*(s')| & \leq \lip \|s - s'\|.
\end{align} 
\end{condition}

\medskip
\noindent{\bf Overview of policy.} 
The policy is exactly the same as that described in Section \ref{ssec:vanilla.policy}, but with a 
single difference: instead of using the Sparse Sampling Monte Carlo Oracle, we replace it with an efficient
version  that exploits Condition \ref{cond:lip.val} as described next. 

\medskip
\noindent{\bf Efficient Sparse Sampling Monte Carlo Oracle.} We describe a modification 
of the Monte Carlo Oracle based on sparse sampling described earlier. As before, our interest is in
obtaining an approximate estimation of $Q^*(s, a)$ for a given $s \in \mS$ and any 
$a \in \mA$ with minimal number of samples of the underlying MDP. To that end, we shall
utilize property \eqref{eq:bounded.increment} of Condition \ref{cond:lyapunov} and 
Condition \ref{cond:lip.val} to propose a modification of the method described
in Section \ref{ssec:vanilla.policy}. For a given parameter $\varepsilon > 0$, define
\begin{align*}
\mS_\varepsilon  & = \{(k_1 \varepsilon, \dots, k_d \varepsilon): k_1,\dots, k_d \in \Z\}.
\end{align*}
For any $s \in \mS = \R^d$, $\exists \tilde{s} \in \mS_\varepsilon$ so that $|s - \tilde{s}|_\infty \leq \varepsilon$ or 
$\|s - \tilde{s}\|_2 \leq \sqrt{d} \|s-\tilde{s}\|_\infty \leq \varepsilon \sqrt{d}$. Define 
$S_\varepsilon: \mS \to \mS_\varepsilon$ that maps $s \in \mS$ to its closest element 
in $\mS_\varepsilon$: i.e. $S_\varepsilon(s) \in \arg\min\{\|s - \tilde{s}\|_2 : \tilde{s} \in \mS_\varepsilon\}$. 

For a given $s \in \mS$, we shall obtain a good estimate of $Q^*(s, a)$ effectively 
using method described in Section \ref{ssec:vanilla.policy}. We start with $s$  as the root node.
For each state-action pair $(s,a),$ $a \in \mA$, we sample next state of 
MDP $C$ times leading to states $\Sn(s, a, C) \subset \mS$. In contrast to the method described in Section \ref{ssec:vanilla.policy}, we use states $\{S_\varepsilon(s'): s' \in \Sn(s, a, C)\}$
 in place of $\Sn(s, a, C)$. These form states (or nodes) as part of the sampling tree at level $1$. For each state, say $\tilde{s}$ at level $1$, for each $a \in \mA$, we sample $C$ next states of MDP, and replace them by their closest elements in $\mS_\varepsilon$ to obtain
 states or nodes at level $2$, we repeat till $H$ levels. Note that all states on level $1$ to $H$ are from $\mS_\varepsilon$. {To improve sample efficiency, during the construction, if a state $\tilde{s}$ has been visited before, we use previously sampled next states $\Sn(\tilde{s},a,C)$ for each action $a\in \mA$, instead of obtaining new samples. }

To obtain an estimate $\hQ^0(s, a)$ for the Q-value associated with the root state $s$ and any action $a \in \mA$, we start by assigning the value $0$ to each leaf node $s_{\sf leaf}$ at depth $H$, i.e., $\hV^H(s_{\sf leaf}) = 0$. These values, 
together with the associated rewards on the edges, are then backed-up to find 
estimates for $Q$-values of their parents at depth $H-1$, and this is repeated till we reach the
root node, $s$. Precisely, for $1 \leq h \leq H$ and node 
$\tilde{s}$ at level $h-1$,
\begin{align}\label{eq:mco.1}
\hQ^{h-1}(\tilde{s},a) & = R_{\tilde{s} a}+ \gamma \frac{1}{C}\sum_{s' \in \Sn(\tilde{s},a, C)} \hV^{h}(S_\varepsilon(s')), 
\qquad \hV^{h-1}(\tilde{s})  = \max_a \hQ^{h-1}(\tilde{s},a). 
\end{align}
The method outputs $\hQ^{0}(s, a)$ as estimate of $Q^*(s, a)$ for all $a \in \mA$. 

\subsubsection{Stability Guarantees}
\noindent{\bf Improved Sample Complexity.} In 
Lemma~\ref{lem:sparse.mod}, we summarize the estimation error as well as the number of samples 
utilized by this modified Sparse Sampling Monte Carlo Oracle. 
\begin{lem}[Modified Oracle Guarantee]
\label{lem:sparse.mod}
Given input state $s \in \mS$, $\delta > 0$, under the modified Sparse Sampling Monte Carlo Oracle, we have 
\begin{align*}
    \max_{a \in \mA} |Q^*(s,a) - \hQ^0(s,a)| &\leq 2 V_{\max} \delta, 
\end{align*}
with probability at least $1-\delta$, with choice of parameters 
$H = \lceil \log_\gamma(\delta) \rceil$,  $\varepsilon =  \frac{\delta V_{\max} (1-\gamma)}{2 \lip \gamma \sqrt{d}}$, and 
\begin{align*}
C & = \Omega\Big(\frac{\gamma^2}{(1-\gamma)^2 \delta^2} \Big(\log \frac{2 | \mA| }{\delta} + d \log H + d \log \frac{1}{\varepsilon}\Big)\Big)~=~\Omega\Big(\frac{1}{\delta^2} \log \frac{1}{\delta}\Big). 
\end{align*}
The number of simulation steps of MDP utilized, as a function of $\delta$ scales as 
$O\Big(\big(\frac{1}{\delta^{2+d}} \log^{1+d} \frac{1}{\delta}\big)\Big)$.

\end{lem}

\medskip
\noindent{\bf Stability.} We state the result establishing stability of the stationary policy described above under appropriate choice of parameters $\delta, \tau, C$ and $H > 0$. 
\begin{thm}
\label{thm:main.2}
Let the MDP of interest satisfy Conditions \ref{cond:1}, \ref{cond:lyapunov} and \ref{cond:lip.val}. Then, with 
$\delta = \tau^2$, $\tau$ as 
\begin{align}\label{cond:tau.2}
\tau & \leq \tau(\alpha) \triangleq  \min\Bigg\{ \sqrt{\frac{\alpha}{24\nu}}, \frac{1-\gamma}{8R_{\max}},
\frac{(1-\gamma)\alpha}{48 R_{\max}|\mA|^2 \nu},  
 \frac{\Delta_{\min}}{\ln \Big(\frac{24\nu |\mA|}{\alpha}\Big)}\Bigg\}, 
\end{align}
and $C, H > 0$ chosen as per Lemma \ref{lem:sparse.mod} for given $\delta > 0$, the resulting stationary 
policy is stable.
\end{thm}
As discussed earlier, as $\alpha \to 0$, with above choice $\delta = \Theta(\alpha^2)$, and hence 
the number of samples required per time step scale as $O\Big(\frac{1}{\alpha^{2d+4}}\log^{d+1} \frac{1}{\alpha}\Big)$. That is, the sample complexity per time step is polynomial in the Lyapunov drift parameter $\alpha$, rather than super-polynomial as required in Theorem~\ref{thm:main.1}.

\subsection{Discovering Appropriate Policy Parameter}

\noindent{\bf The Purpose.}  The sample inefficient policy described in Section \ref{ssec:vanilla.policy} or sample efficient policy described in 
Section \ref{ssec:efficient.policy} are stable only if the parameter $\delta$ (or equivalently $\tau$, since $\delta = \tau^2$ in Theorem 
\ref{thm:main.1} or \ref{thm:main.2}) is chosen to be small enough. However, what is small enough value for $\delta$ is not clear a priori 
without knowledge of the system parameters as stated in \eqref{cond:tau} or \eqref{cond:tau.2}. In principle, we can continue reducing 
$\delta$; however, the sample complexity required per unit time increases with reduction in $\delta$. Therefore, it is essential to ensure that the reduction is eventually terminated. 
Below we describe such a method, based on a hypothesis test using the positive recurrence property established in the proofs of Theorems \ref{thm:main.1} and \ref{thm:main.2}.

\medskip
\noindent{\bf A Statistical Hypothesis Test.}  Toward the goal of finding an appropriate value of $\delta$, i.e., small enough but not too small, we describe a statistical hypothesis test --- if $\delta$ is small enough, then test passes with high probability. We shall utilize this test to devise an adaptive method that finds the right value of $\delta$ as described below. We state the following structural assumption. 
\begin{condition}\label{cond:lyap.lb}
Consider the setting of Condition \ref{cond:lyapunov}. Let the Lyapunov function $L:\mS\rightarrow\R_{+}$, in 
addition, satisfy 
\begin{align}\label{eq:lyap.lb}
L(s) & \geq c_1 \|s\|_2 + c_2, 
\end{align}
for all $s \in \mS$ with constants $c_1, c_2 \in \R$ and $c_1 > 0$. 
\end{condition}
Let $\tau(\alpha)$ be as defined in \eqref{cond:tau} and  $\delta(\alpha) = \tau(\alpha)^2$.  
Under the above condition, arguments used in the proofs of Theorem~\ref{thm:main.1} and~\ref{thm:main.2} establish that if $\delta \le \delta(\alpha)$, then the following exponential tail bound holds:
\begin{align*}
\P\left(\left\Vert s_{t}\right\Vert \ge b \right) & \le C e^{-\eta c_{1} b},\quad \forall b>0, t \ge 1.
\end{align*}
Note that the probability bound on the right hand side is summable over $t$ when choosing, say, $b=\log^2 t$.
This property enables a hypothesis test, via checking $\|s_t\|$, which is used below to devise an adaptive method.
The norm in use can be arbitrary since all norms  are equivalent in a finite-dimensional vector space.

\medskip
\noindent{\bf An Adaptive Method for Tuning $\delta$.} Using the statistical hypothesis test, we now describe
 an algorithm that finds $\delta$ under which the hypothesis test is satisfied with probability exponentially close to $1$, and 
$\delta$ is strictly positive.  Initially, set $\delta_0 = 1$. 
At each time $t$, we decide to whether adjust value of $\delta$ or not by checking whether $\left\Vert s_{t}\right\Vert \ge\log^{2}t$. 
If yes, then we set $\delta_{t+1} = \delta_t/2$; else we keep $\delta_{t+1} = \delta_t$. 

\subsubsection{Stability Guarantees}
The following theorem provides guarantees for the above adaptive algorithm.

\begin{thm}\label{thm:main.3}
Consider the setup of Theorem~\ref{thm:main.1} (respectively Theorem~\ref{thm:main.2}). Let, in
addition, Condition~\ref{cond:lyap.lb} hold. Consider the system operating with choice
of parameter $\delta=\tau^{2}$ at any time as per the above described
method with policy described in Section~\ref{ssec:vanilla.policy} (respectively Section~\ref{ssec:efficient.policy}). Then with probability 1, the system operating under such changing
choice of $\delta$ is either eventually operating with value $\delta\le\delta(\alpha)$
and hence stable, or 
\begin{equation}
\lim\sup_{t\to\infty}\frac{\left\Vert s_{t}\right\Vert }{\log^{2}t}=O(1);\label{eq:limsup_bound}
\end{equation}
moreover, we have 
\begin{equation}
\lim\inf_{t\to\infty}\delta_{t}>0.\label{eq:delta_bound}
\end{equation}
\end{thm}

The theorem guarantees that the system is either stable as the algorithm finds the appropriate policy parameter, or near-stable in the sense 
that the state grows at most as $\log^2 t$. That is, this adaptive algorithm induces at worst $O(\log^2 t)$ regret since the optimal
policy will retain $\|s_t\| = O(1)$. This is, indeed a low-regret algorithm in that respect. 

\medskip
\noindent{\bf Connecting to Fluid Stability.} When viewed in the context of queueing system, this result suggests that the queue-sizes (which is the state 
of the system) scales as $O(\log^2 t)$. As mentioned earlier, this is related to fluid stability. Precisely, consider re-scaling  
$\bar{s}^n_t = \frac{1}{n} s_{nt}$. Then, by definition, we have $\|\bar{s}^n_t \|_2 = O\big( \frac{(\log n + \log t)^2 } {n} \big) $. For any fixed 
$t \geq 0$, this would suggest $\|\bar{s}^n_t \|_2 \to 0$ as $n\to \infty$. Since the system trajectory of state $s_t, t \geq 0$ induces Lipschitz
sample paths, for any $t \in [0,T]$ the limit points of $\|\bar{s}^n_t \|_2$ exist (due to compactness of appropriately defined metric space)
and they are all ${\bf 0}$. This will imply fluid or rate stability in a queueing system: the ``departure rate'' of jobs is the same 
as the ``arrival rate'' of jobs, cf.~\cite{dai1995unified,dai1995stability}. This is a highly desirable guarantee for queueing systems 
that is implied by our sample efficient, online and adaptive RL policy.

\section{Conclusion}
\label{sec:conclusion}

In this paper, we investigate  reinforcement learning for systems with unbounded state space, as motivated by classical queueing systems. To tackle the challenges due to unboundedness of the state space, we propose a natural notion of stability to quantify the ``goodness'' of policy and importantly, design efficient, adaptive algorithms that achieves such stability.

Stability in problems with unbounded state spaces are of central importance in classical queueing and control theory, yet this aspect has received relatively less attention in existing reinforcement learning literature. As reinforcement learning becomes increasingly popular in various application domains, we believe that modeling and achieving stability is critical. This paper and the framework introduced provide some first steps towards this direction.

\bibliographystyle{plain}
\bibliography{stable}

\appendix
\appendixpage
\section{ Related Work}
\label{app:related}
The concept of stability introduced in this paper is related to the notion 
of \emph{safety} in RL, but with crucial differences. Various definitions of safety exist in literature~\cite{pecka2014safe,garcia2015comprehensive}. One line of work defines safety as 
hard constraints on individual states and/or actions~\cite{garcia2012safe,hans2008safe,dalal2018safe,dean2019safely,koller2018learning}. 
Some other work considers safety guarantee in terms of keeping the expected sum of certain costs 
over a trajectory below a given threshold~\cite{achiam2017constrained,chow2018lyapunov,yu2019convergent}. 
In our work, stability is defined in terms of the positive recurrence of Markov chains, which cannot be immediately written as constraints/cost over the state and actions. In particular, our stability notion captures long-term behaviors of the system---it should eventually stay in a desirable region of the state space with high probability. In general, there does not exist an action that immediately drives the system back to that region; learning a policy that achieves so in the long run is non-trivial and precisely our goal.

Many work on RL safety is model-based, either requiring  a prior known safe backup policy~\cite{garcia2012safe,hans2008safe}, or using model predictive control approaches~\cite{wabersich2018safe,sadigh2016safe,aswani2013provably,koller2018learning}. 
One line of work focuses specifically on systems with a linear model with constraints (e.g., LQR)~\cite{chen2018approximating,dean2019safely}. Some other work considers 
model-free policy search algorithms~\cite{achiam2017constrained,chow2018lyapunov,yu2019convergent}, 
under the framework of constrained Markov decision process~\cite{altman1999constrained}, 
which models safety as expected cumulative constraint costs.

Another line of work considers control-theoretic notions of stability~\cite{berkenkamp2016safe,vinogradska2016stability,berkenkamp2017safe}, which bears similarity to our framework. We remark that these results mostly focus on systems with deterministic and partially unknown dynamics, different from our setting where the dynamics are stochastic and unknown. Their approaches are limited to compact state spaces where discretization is feasible.

Our analysis makes use of Lyapunov functions, which is a classical tool in control and Markov chain theory for studying stability and steady-state behaviors~\cite{glynn2008bounding,eryilmaz2012asymptotically}. The work~\cite{perkins2002lyapunov} is among the first to use Lyapunov functions in RL and studies closed-loop stability of an agent. More recent work uses Lyapunov functions to establish the finite-time error bounds of TD-learning ~\cite{srikant2019finite} and to solve constrained MDPs~\cite{chow2018lyapunov} and to find region of attraction for deterministic systems~\cite{berkenkamp2017safe,berkenkamp2016safe}.

Our RL algorithm fits broadly into value-based methods~\cite{watkins1992q,sutton1988learning,mnih2015human,van2016deep,yang2019harnessing,shah2018qlnn}. Approximate dynamic programming techniques and RL have been applied to queueing problems in prior work~\cite{krishnasamy2018learning,roy2003approximate,moallemi2008approximate}, though their settings and goals are quite different from us, and their approaches exploit prior knowledge of queueing theory and specific structures of the problems. Most related to us is the recent work~\cite{liu2019reinforcement}, which also considers problems with unbounded state spaces. Their algorithm  makes use of a \emph{known} stabilizing policy. We do not assume knowledge of such a policy; rather, our goal is to learn one from data.

\section{Proof of Theorem \ref{thm:main.1}}
\label{sec:analysis}
For any given $\delta > 0$, we know that at each step, with 
probability $1-\delta$,
\begin{align}\label{eq:q-err}
    |Q^*(s_t,a) - \hat{Q}(s_t,a)|
    & \leq \epsilon \triangleq  \frac{2R_{\max}}{1-\gamma}\delta,\quad \forall a\in \mA,
\end{align}
with appropriate choice of parameters $C, H$ as stated in Lemma \ref{lem:sparse}.
The stationary policy utilizes Botzman transformation of $\hat{Q}$. The following lemma
establishes the approximation error between the Boltzman policy and optimal policy.

\begin{lem}
\label{lem:boltzmann}
Given state $s\in\mS,$ let $\hat{Q}(s,a)$ be such that 
with probability at least $1-\delta$, 
\[
|\hat{Q}(s,a)-Q^{*}(s,a)|\leq\epsilon,\forall \:a\in\mA.
\]
Consider two Boltzmann policies with temperature $\tau:$
\[
\hat{P}(s,a)=\frac{e^{\hat{Q}(s,a)/\tau}}{\sum_{a'}e^{\hat{Q}(s,a')/\tau}} \textrm{~~~~and~~~~} P^{*}(s,a)=\frac{e^{Q^{*}(s,a)/\tau}}{\sum_{a'}e^{Q^{*}(s,a')/\tau}},\quad \forall \:a\in\mA.
\]
Then, we have that 
\begin{enumerate}
\item With probability at least $1-\delta,$
\[
\left\Vert \hat{P}(s)-P^{*}(s)\right\Vert _{\textup{TV}}\leq\frac{|\mA|^{2}}{2}\cdot\frac{e^{2\epsilon/\tau}-1}{e^{2\epsilon/\tau}+1} .
\]
\item With probability at least $1-\delta$,
\begin{align*}
\left\Vert \hat{P}(s)-\pi^{*}(s)\right\Vert _{\textup{TV}}\leq\frac{|\mA|^{2}}{2}\cdot\frac{e^{2\epsilon/\tau}-1}{e^{2\epsilon/\tau}+1}+\big(|\mA|-1\big)e^{-\frac{\Delta_{\min}(s)}{\tau}}. 
\end{align*}
\end{enumerate}
\end{lem}
From above Lemma, with notation $\Delta_{\min} \leq \Delta_{\min}(s)$ for any $s\in\mS$, 
we obtain that for our stochastic policy $\hat{P}$, with probability $1-\delta$,
\begin{align}\label{eq:approx.policy}
\left\Vert \hat{P}(s_t)-\pi^{*}(s_t)\right\Vert _{\text{TV}}
& \leq \kappa 
\triangleq \frac{|\mA|^{2}}{2}\cdot\frac{e^{2\epsilon/\tau}-1}{e^{2\epsilon/\tau}+1}+\big(|\mA|-1\big)e^{-\frac{\Delta_{\min}}{\tau}} .
\end{align}
By Condition \ref{cond:lyapunov}, we know that the MDP dynamic under the optimal policy respects a Lyapunov function that has bounded increment and drift property. As per \eqref{eq:approx.policy}, the Boltzman policy is
a good approximation of the optimal policy at each time step with high probability. The following Lemma
argues that under such an approximate policy, MDP respects the same Lyapunov function but with
a slightly modified drift condition. 

\begin{lem}
\label{lem:pi_drift}
Consider the set up of Theorem \ref{thm:main.1}. Suppose that at each time step $t$, a stochastic policy $\pi(s_t)$ (i.e., $\pi(s_t)$ is a 
distribution over $\mathcal{A}$) is executed such that for each $t$, with probability 
at least $1-\delta$, 
\[
\left\Vert \pi(s_{t})-\pi^{*}(s_{t})\right\Vert _{\textup{TV}}\leq\kappa.
\]
Then, for every $s\in \mS$ such that $L(s)> B$, we have
\begin{equation*}
    \E[L(s_{t+1}) - L(s_t)|s_t = s] \leq 4\nu\big((1-\delta)\kappa + \delta\big)-\alpha. 
\end{equation*}
\end{lem}
Now, based on Lemma \ref{lem:pi_drift}, we note that for every $s\in\mS$ such that $L(s)>B$, 
the following drift inequality holds for our stochastic policy $\hat{P}$:
\begin{align}
\label{eq:proof:thm:drift:hatpi}
    \E[L(s_{t+1}) - L(s_t)|s_t = s] & \leq 4\nu\big((1-\delta)\kappa + \delta\big)-\alpha. 
\end{align}
We shall argue that under choice of $\delta = \tau^2$ and $\tau$ as per \eqref{cond:tau}, the right hand 
size of \eqref{eq:proof:thm:drift:hatpi} is less than $-\frac{1}{2}\alpha$. To do so, it is sufficient to argue
that $4\nu(\kappa + \delta)\leq \alpha/2$. That is, we want to argue
\begin{align}
\label{eq:thm:proof:negative_suffcient}
    4\nu\Big(\underbrace{\frac{|\mA|^{2}}{2}\cdot\big(1-\frac{2}{e^{4R_{\max}\delta/(\tau(1-\gamma))}+1}\big)}_\text{(I)}+\underbrace{\big(|\mA|-1\big)e^{-\frac{\Delta_{\min}}{\tau}}}_\text{(II)} + \underbrace{\delta}_\text{(III)}\Big) & \leq \frac{1}{2}\alpha. 
\end{align} 
To establish the above mentioned claim, it is sufficient to argue that each of (I), (II) and (III) is no more than 
$\alpha/24\nu$, under the choice of $\tau$ as per \eqref{cond:tau} and $\delta=\tau^2$. To that end, (III) is 
less than $\alpha/24\nu$ immediately since $\tau \leq \sqrt{\frac{\alpha}{24\nu}}$. 
For (II), similar claim follows due to $\tau \leq \frac{\Delta_{\min}}{\ln \Big(\frac{24\nu |\mA|}{\alpha}\Big)}$. For (I), 
using facts that $e^x \leq 1+2x, ~x \in (0,1)$ and $1/(1+x) > 1-x$ for $x \in (0,1)$ and 
$4R_{\max}\delta/(\tau(1-\gamma)) = 4R_{\max}\tau/(1-\gamma) \leq \frac{1}{2}$, we have that (with $\delta = \tau^2$)
\begin{align*}
\frac{|\mA|^{2}}{2}\cdot\big(1-\frac{2}{e^{4R_{\max}\delta/(\tau(1-\gamma))}+1}\big) & \leq 
\frac{|\mA|^{2}}{2}\cdot\big(1-\frac{2}{8R_{\max}\tau/(1-\gamma) + 2}\big) \\
&\leq \frac{|\mA|^{2}}{2}\cdot\big(4R_{\max}\tau/(1-\gamma)\big) 
\leq \frac{\alpha}{24 \nu}.
\end{align*}
Thus, we conclude that if $L(s) > B$, then 
\begin{align}\label{eq:fin.drift}
    \E[L(s_{t+1}) - L(s_t)|s_t = s] & \leq -\frac{1}{2} \alpha.
\end{align}
We recall the following result of \cite{hajek1982hitting} that implies positive recurrence property for stochastic system satisfying
drift condition as in \eqref{eq:fin.drift}. 
\begin{lem}
\label{lem:hajek}
Consider a policy $\pi$. Suppose that there exists a Lyapunov function $L$ such that the policy $\pi$ satisfies the bounded increment condition with parameter $\nu>0$ and the drift condition with parameters $\alpha>0$ and $B>0$. Let $T_a\triangleq \min\{t\geq 0 :L(s_t)\leq a\}$. Let $c(\nu) = e^\nu - \nu - 1$, $\eta = \min(1,\alpha/2c(\nu))$
and $\rho = 1- \eta\alpha/2c(\nu) < 1$. Then it follows that for all $b \geq 0$, 
\begin{align}
    \P(L(s_t)\geq b|s_0=s)&\leq \rho^te^{\eta(L(s)-b)} + \frac{1-\rho^t}{1-\rho}e^{\nu + \eta(B - b)}, \label{eq:lem:hajek}\\
    \P(T_a > k|s_0=s)&\leq e^{\eta(L(s)-a)}\rho^k, \quad\forall\: a\geq B.  \label{eq:lem:hajek:recurrent}
\end{align}
\end{lem}
By immediate application of Lemma \ref{lem:hajek}, with $c = c(\nu) = e^\nu - \nu - 1$, 
$\eta = \min(1,\alpha/4c(\nu))$ and $\rho = 1- \eta\alpha/4c(\nu) < 1$, it follows that 
\begin{align}
    \P\big(L(s_t) \geq b | s_0=s \big) &\leq {\rho}^te^{{\eta}(L(s)- b) } + \frac{1-{\rho}^t}{1-{\rho}}e^{{\eta}({B}  - b) + {\nu}},\quad\forall\:s\in\mS, \label{eq:thm:stability}
\end{align}
and $\forall\:b \geq B$,
\begin{align}
    \P\big(T_b(t) > k \:|\: s_t=s\big) & \leq e^{{\eta}(L(s)- b)}\rho^k,\quad\forall\:s\in\mS, \label{eq:thm:recurrent}
\end{align}
where $T_b(t) \triangleq \min\{k\geq 0\::\:L(s_{t+k})\leq b\}$ is the return time to a set that the Lyapunov function is bounded by $b$ starting from time $t$.

Define level set $\mathcal{C}(\ell )\triangleq \{s\in\mS\::\: L(s) \leq \ell \}$. By Condition \ref{cond:lyapunov}, 
$\sup_{s\in\mathcal{C}(\ell)}\Vert s\Vert_2 <\infty$ for any finite $\ell $. 
Now \eqref{eq:thm:stability} implies that for any small $\phi>0$, we have
\begin{equation*}
    \lim\sup_{t\rightarrow\infty}\P\big(s_t\notin \mathcal{C}(b+\phi)|s_0=s\big)\leq \lim\sup_{t\rightarrow\infty}\P\big(L(s_t)\geq b|s_0=s\big)\leq \frac{1}{1-\rho}e^{\eta(B-b) +\nu}.
\end{equation*}
By letting $b\geq B$ and $b$ be large enough, we can always make the above probability bound as small as possible. That is, for any given  $\theta > 0 $, there exist a large $b\geq B$ and a small $\phi >0$, such that
\begin{equation*}
    \lim_{t\rightarrow\infty}\P\big(s_t\notin \mathcal{C}(b+\phi)|s_0=s\big)\leq \theta,\quad\forall\:s\in\mS.
\end{equation*}
In addition, \eqref{eq:thm:recurrent} implies that 
\begin{equation*}
    \P\big(T_{b+\phi}(t) > k \:|\: s_t=s\big)\leq e^{{\eta}\big(L(s)- b-\phi \big)}\rho^k,\quad\forall\:s\in\mS\textrm{ and }\forall\:t\geq 0.
\end{equation*}
Therefore, 
\begin{equation*}
    \E\big[T_{b+\phi}(t)|s_t = s\big]=\sum_{k = 0}^\infty\P\big(T_{b+\phi}(t) > k|s_t = s\big)\leq e^{{\eta}(L(s)- b-\phi)}\cdot\frac{1}{1-\rho}<\infty.
\end{equation*}
That is, given the current state $s$, the return time to the bounded set $\mathcal{C}(b+\phi)$ is 
uniformly bounded, across all $t$. This establishes the stability of the policy as desired and Theorem \ref{thm:main.1}.

\subsection{Proof of Lemma \ref{lem:boltzmann}}

We first bound the total variation distance between $\hat{P}$ and
$P^{*}$.  
For each $a$, we have 
\begin{align*}
\left|\hat{P}(s,a)-P^{*}(s,a)\right| & =\left|\frac{e^{\hat{Q}(s,a)/\tau}}{\sum_{a'}e^{\hat{Q}(s,a')/\tau}}-\frac{e^{Q^{*}(s,a)/\tau}}{\sum_{a^{''}}e^{Q^{*}(s,a^{''})/\tau}}\right|\\
 & =\frac{\left|e^{\hat{Q}(s,a)/\tau}\sum_{b}e^{Q^{*}(s,b)/\tau}-e^{Q^{*}(s,a)/\tau}\sum_{b}e^{\hat{Q}(s,b)/\tau}\right|}{\left(\sum_{a'}e^{\hat{Q}(s,a')/\tau}\right)\left(\sum_{a^{''}}e^{Q^{*}(s,a^{''})/\tau}\right)}\\
 & = \sum_{b}\frac{\left|e^{\hat{Q}(s,a)/\tau+Q^{*}(s,b)/\tau}-e^{Q^{*}(s,a)/\tau+\hat{Q}(s,b)/\tau}\right|}{\left(\sum_{a'}e^{\hat{Q}(s,a')/\tau}\right)\left(\sum_{a^{''}}e^{Q^{*}(s,a^{''})/\tau}\right)}.
\end{align*}
Consider $b$-th term in the above summation:
\begin{align*}
& \quad \frac{\left|e^{\hat{Q}(s,a)/\tau+Q^{*}(s,b)/\tau}-e^{Q^{*}(s,a)/\tau+\hat{Q}(s,b)/\tau}\right|}{\left(\sum_{a'}e^{\hat{Q}(s,a')/\tau}\right)\left(\sum_{a^{''}}e^{Q^{*}(s,a^{''})/\tau}\right)} 
 \le\frac{\left|e^{\hat{Q}(s,a)/\tau+Q^{*}(s,b)/\tau}-e^{Q^{*}(s,a)/\tau+\hat{Q}(s,b)/\tau}\right|}{e^{\hat{Q}(s,a)/\tau+Q^{*}(s,b)/\tau}+e^{Q^{*}(s,a)/\tau+\hat{Q}(s,b)/\tau}}\\
 & \le\frac{e^{Q^{*}(s,a)/\tau+\hat{Q}(s,b)/\tau}\cdot\left|e^{\left[\hat{Q}(s,a)+Q^{*}(s,b)-Q^{*}(s,a)-\hat{Q}(s,b)\right]/\tau}-1\right|}{e^{Q^{*}(s,a)/\tau+\hat{Q}(s,b)/\tau}\cdot\left(e^{\left[\hat{Q}(s,a)+Q^{*}(s,b)-Q^{*}(s,a)-\hat{Q}(s,b)\right]/\tau}+1\right)}\\
 & =\frac{\left|e^{\left[\hat{Q}(s,a)+Q^{*}(s,b)-Q^{*}(s,a)-\hat{Q}(s,b)\right]/\tau}-1\right|}{e^{\left[\hat{Q}(s,a)+Q^{*}(s,b)-Q^{*}(s,a)-\hat{Q}(s,b)\right]/\tau}+1}
 =\frac{\left|e^{t}-1\right|}{e^{t}+1},
\end{align*}
where $t :=\left[\hat{Q}(s,a)+Q^{*}(s,b)-Q^{*}(s,a)-\hat{Q}(s,b)\right]/\tau$. By assumption, we have $\left|t\right|\le2\epsilon/\tau$. Now, 
if $t > 0$, then 
\[
\frac{\left|e^{t}-1\right|}{e^{t}+1}=\frac{e^{t}-1}{e^{t}+1}=1-\frac{2}{e^{t}+1}\le1-\frac{2}{e^{2\epsilon/\tau}+1}=\frac{e^{2\epsilon/\tau}-1}{e^{2\epsilon/\tau}+1}.
\]
Similarly, if $t < 0$, then 
\[
\frac{\left|e^{t}-1\right|}{e^{t}+1}=\frac{-e^{t}+1}{e^{t}+1}=-1+\frac{2}{e^{t}+1}\le-1+\frac{2}{e^{-2\epsilon/\tau}+1}=\frac{-e^{-2\epsilon/\tau}+1}{e^{-2\epsilon/\tau}+1}=\frac{-1+e^{2\epsilon/\tau}}{e^{2\epsilon/\tau}+1}.
\]
That is, for any $\left|t\right|\le2\epsilon/\tau$, we have 
$\frac{\left|e^{t}-1\right|}{e^{t}+1}\le\frac{e^{2\epsilon/\tau}-1}{e^{2\epsilon/\tau}+1}$.
Combining the above results, we obtain that 
\begin{align*}
\left|\hat{P}(s,a)-P^{*}(s,a)\right|\le\sum_{b}\frac{e^{2\epsilon/\tau}-1}{e^{2\epsilon/\tau}+1}\le & |\mA|\cdot\frac{e^{2\epsilon/\tau}-1}{e^{2\epsilon/\tau}+1}.
\end{align*}
Therefore, we have
\[
\left\Vert \hat{P}(s)-P^{*}(s)\right\Vert _{\text{TV}}=\frac{1}{2}\sum_{a\in\mA}\left|\hat{P}(s,a)-P^{*}(s,a)\right|\leq\frac{|\mA|^{2}}{2}\cdot\frac{e^{2\epsilon/\tau}-1}{e^{2\epsilon/\tau}+1}.
\]
Next, we bound $\left\Vert P^{*}(s)-\pi^{*}(s)\right\Vert _{\text{TV}}$.
Let $a^{*}\in\mA^{*}(s)$. Notice that for each $a\in\mA^{*}(s),$
$P^{*}(s,a)=P^{*}(s,a^{*})\leq\frac{1}{|\mA^{*}(s)|}$. For each $a\notin\mA^{*}(s)$,
we have 
\[
P^{*}(s,a)=\frac{e^{Q^{*}(s,a)/\tau}}{\sum_{a'}e^{Q^{*}(s,a')/\tau}}\leq\frac{e^{Q^{*}(s,a)/\tau}}{e^{Q^{*}(s,a^{*})/\tau}}=e^{-\frac{Q^{*}(s,a^{*})-Q^{*}(s,a)}{\tau}}\leq e^{-\Delta_{\min}(s)/\tau}.
\]
Note that for each $a\in\mA^{*}(s),$ $\pi^{*}(a | s)=\frac{1}{|\mA^{*}(s)|}$,
and for each $a\notin\mA^{*}(s),$ $\pi^{*}(a | s)=0$. It hence follows
that 
\begin{align*}
\left\Vert P^{*}(s)-\pi^{*}(s)\right\Vert _{\text{TV}} & =\frac{1}{2}\sum_{a\in\mA}|P^{*}(s,a)-\pi^{*}(a | s)|\\
 & =\frac{1}{2}\sum_{a\in\mA^{*}(s)}\big(\pi^{*}(a | s)-P^{*}(s,a)\big)+\frac{1}{2}\sum_{a\notin\mA^{*}(s)}P^{*}(s,a)\\
 & =\sum_{a\notin\mA^{*}(s)}P^{*}(s,a)
 \leq\sum_{a\notin\mA^{*}(s)}e^{-\Delta_{\min}(s)/\tau}\\
 & \leq(|\mA|-1)e^{-\Delta_{\min}(s)/\tau}.
\end{align*}
By triangle inequality, we have
\begin{align*}
\left\Vert \hat{P}(s)-\pi^{*}(s)\right\Vert _{\text{TV}} & \leq\left\Vert \hat{P}(s)-P^{*}(s)\right\Vert _{\text{TV}}+\left\Vert P^{*}(s)-\pi^{*}(s)\right\Vert _{\text{TV}}\\
 & \leq\frac{|\mA|^{2}}{2}\cdot\frac{e^{2\epsilon/\tau}-1}{e^{2\epsilon/\tau}+1}+(|\mA|-1)e^{-\Delta_{\min}(s)/\tau}.
\end{align*}
This concludes the proof of Lemma \ref{lem:boltzmann}.

\subsection{Proof of Lemma \ref{lem:pi_drift}}

By Condition \ref{cond:lyapunov}, for each $s_{t}\in\mS$ we have
\[
|L(s_{t+1})-L(s_{t})|\leq\nu,
\]
where $s_{t+1}\sim p(\cdot|s,\pi(s_{t}))$. 
Let us analyze the drift of $L$ under the stochastic policy $\pi(s_t)$.
Fix a state $s\in\mS$ such that $L(s)> B$. Then,
\begin{align}
&\E[L(s_{t+1})-L(s_{t})|s_{t}=s] = 
\Big(\sum_{s'\in\mS}L(s')\sum_{a\in\mA}\pi(a | s) p(s'|s,a)\Big)-L(s)\nonumber \\
=& \Big(\sum_{s'\in\mS}L(s')\sum_{a\in\mA}\big(\pi(a | s)-\pi^{*}(a | s)\big)p(s'|s,a)+\sum_{s'\in\mS}L(s')\sum_{a\in\mA}\pi^{*}(a | s)p(s'|s,a)\Big)-L(s)\nonumber \\
=& \sum_{a\in\mA}\big(\pi(a | s)-\pi^{*}(a | s)\big)\sum_{s'\in\mS}L(s')p(s'|s,a)+\E_{\pi^{*}}[L(s_{t+1})-L(s_{t})|s_{t}=s]. \label{eq:drift_appro_policy-1}
\end{align}
To analyze the first term of \eqref{eq:drift_appro_policy-1}, we define the sets
\begin{equation*}
    \mA_{\pi}^{+}=\{a\in\mA\:|\:\pi(a | s)\geq\pi^{*}(a | s)\}
\textrm{~~~~and~~~~} \mA_{\pi}^{-}=\mA\backslash\mA_{\pi}^{+}.
\end{equation*} Note that $\sum_{a\in\mA}\pi(a | s)=\sum_{a\in\mA}\pi^{*}(a | s)=1$.
Therefore, we have
\begin{align}
&\sum_{a\in\mA}\big(\pi(a | s)-\pi^{*}(a | s)\big)\sum_{s'\in\mS}L(s')p(s'|s,a) \nonumber\\
=& \sum_{a\in\mA}\big(\pi(a | s)-\pi^{*}(a | s)\big)\sum_{s'\in\mS}L(s')p(s'|s,a)+\sum_{a\in\mA}\big(\pi(a | s)-\pi^{*}(a | s)\big)L(s)\nonumber \\
 =& \sum_{a\in\mA_{\pi}^{+}}\big(\pi(a | s)-\pi^{*}(a | s)\big)\Big[\sum_{s'\in\mS}L(s')p(s'|s,a)-L(s)\Big]\label{eq:drift_positive_part-1}\\
 & \quad+\sum_{a\in\mA_{\pi}^{-}}\big(\pi(a | s)-\pi^{*}(a | s)\big)\Big[\sum_{s'\in\mS}L(s')p(s'|s,a)-L(s)\Big].\label{eq:drift_negative_part-1}
\end{align}
For the term (\ref{eq:drift_positive_part-1}), note that for each $a$ and $s'$
such that $p(s'|s,a)>0$, Condition~\ref{cond:lyapunov} ensures that
\[
|L(s')-L(s)|\leq\nu.
\]
Consequently, we have
\begin{align*}
(\ref{eq:drift_positive_part-1}) & \leq\sum_{a\in\mA_{\pi}^{+}}\big(\pi(a | s)-\pi^{*}(a | s)\big)\Big[\sum_{s'\in\mS}\big(L(s)+\nu\big)p(s'|s,a)-L(s)\Big]\\
 & =\sum_{a\in\mA_{\pi}^{+}}\big(\pi(a | s)-\pi^{*}(a | s)\big)\nu
 ~\leq2\nu\Vert\pi(s)-\pi^{*}(s)\Vert_{\text{TV}}.
\end{align*}
Following similar argument, 
\begin{align*}
(\ref{eq:drift_negative_part-1}) 
 & \leq2\nu\Vert\pi(s)-\pi^{*}(s)\Vert_{\text{TV}}.
\end{align*}
Combining the above two inequalities, we have
\begin{align}
\sum_{a\in\mA}\big(\pi(a | s)-\pi^{*}(a | s)\big)\sum_{s'\in\mS}L(s')p(s'|s,a) & \leq4\nu\Vert\pi(s)-\pi^{*}(s)\Vert_{\text{TV}}. \label{eq:drift_policy_difference}
\end{align}
Recall that the total variation distance is bounded by $1$ and that the stochastic policy $\pi(s)$ satisfies that with probability at least $1-\delta$, $\Vert\pi(s)-\pi^{*}(s)\Vert_{\text{TV}}\leq \kappa$. 
Taking expectation on both sides of \eqref{eq:drift_policy_difference} with respect to the randomness in the stochastic policy $\pi(s)$, 
we have
\begin{equation}
  \E\Big[\sum_{a\in\mA}\big(\pi(a | s)-\pi^{*}(a | s)\big)\sum_{s'\in\mS}L(s')p(s'|s,a)\Big]  \leq4\nu\big((1-\delta)\kappa +\delta\big).\label{eq:drift_appro_policy_part_1-1}
\end{equation}
Substituting the upper bound (\ref{eq:drift_appro_policy_part_1-1}) into
(\ref{eq:drift_appro_policy-1}) yields 
\[
\E[L(s_{t+1})-L(s_{t})|s_{t}=s]\leq4\nu\big((1-\delta)\kappa +\delta\big)+\E_{\pi^{*}}[L(s_{t+1})-L(s_{t})|s_{t}=s].
\]
Note that by Condition \ref{cond:lyapunov}, for each $s\in\mS$ such that $L(s)> B,$ $\E_{\pi^{*}}[L(s_{t+1})-L(s_{t})|s_{t}=s]\leq-\alpha$.
Finally, we conclude that 
\[
\E_{\pi}[L(s_{t+1})-L(s_{t})|s_{t}=s]\leq4\nu\big((1-\delta)\kappa+\delta\big)-\alpha,
\]
thereby completing the proof.

\subsection{Proof of Lemma~\ref{lem:hajek}}\label{sec:proof_drift}

Throughout the proof, we fix an $\eta\in(0, \min\{1, \alpha/(2c)\})$ and let $\rho = 1 - \eta\alpha/2$. To simplify the notation, we fix a $s_0$, and the probabilities and expectations should be understood as conditioned on $s_0 = s$ when appropriate.
Let $\mF_t$ denote the smallest 
$\sigma$-algebra containing all information pertaining to the MDP up to time $t$, i.e., $\{s_t\}_{t\geq 0}$ is adapted to $\{\mF_t\}_{t\geq0}$. 

We start by establishing \eqref{eq:lem:hajek}. To do so, we will instead prove the following inequality, 
from which \eqref{eq:lem:hajek} can be readily obtained via Markov's inequality:
\begin{equation}
    \E[e^{\eta L(s_t)}]\leq \rho^te^{\eta L(s_0)} + \frac{1-\rho^t}{1-\rho}e^{\nu + \eta B}.\label{eq:lem:hajek:proof1}
\end{equation}
Note that  $\E[e^{\eta L(s_{t+1)}}] = \E[\E[e^{\eta L(s_{t+1})}|\mF_t]]$. Further,
\begin{equation}
    \E[e^{\eta L(s_{t+1})}|\mF_t] = \E[e^{\eta L(s_{t+1})}\mathbb{I}\{L(s_t)\leq B\}|\mF_t] + \E[e^{\eta L(s_{t+1})}\mathbb{I}\{L(s_t)> B\}|\mF_t]. \label{eq:lemma:hajek:proof2}
\end{equation}
We now analyze the two terms on the R.H.S. of the \eqref{eq:lemma:hajek:proof2} separately. For the first term,
\begin{align*}
    \E[e^{\eta L(s_{t+1})}\mathbb{I}\{L(s_t)\leq B\}|\mF_t] & = \E[e^{\eta (L(s_{t+1}) - L(s_{t}))}e^ {\eta L(s_t)}\mathbb{I}\{L(s_t)\leq B\}|\mF_t]\leq e^{\nu + \eta B},
\end{align*}
where the last inequality follows from the bounded increment condition and the fact that $\eta < 1$. For the second term of \eqref{eq:lemma:hajek:proof2}), observe
\begin{align*}
    \E[e^{\eta L(s_{t+1})}\mathbb{I}\{L(s_t)> B\}|\mF_t] & = \E[e^{\eta (L(s_{t+1}) - L(s_{t}))}e^ {\eta L(s_t)}\mathbb{I}\{L(s_t)> B\}|\mF_t]\\
    &\leq \E[e^{\eta (L(s_{t+1}) - L(s_{t}))\mathbb{I}\{L(s_t)> B\}}|\mF_t] \cdot e^ {\eta L(s_t)}.
\end{align*}
We now show that $\E[e^{\eta (L(s_{t+1}) - L(s_{t}))\mathbb{I}\{L(s_t)> B\}}|\mF_t] \leq \rho$. Since $|L(s_{t+1}) - L(s_{t})| \leq \nu$ 
with probability $1$,  $\E[e^{\eta (L(s_{t+1}) - L(s_{t}))\mathbb{I}\{L(s_t)> B\}}|\mF_t]$ has an absolutely convergent series 
expansion. That is,
\begin{align*}
  \E[e^{\eta (L(s_{t+1}) - L(s_{t}))\mathbb{I}\{L(s_t)> B\}}|\mF_t] &= 1 + \eta  \E[{(L(s_{t+1}) - L(s_{t}))\mathbb{I}\{L(s_t)> B\}}|\mF_t] \\
  & + \eta^2\sum_{k=2}^\infty\frac{\eta^{k-2}}{k!} \E[\big((L(s_{t+1}) - L(s_{t}))\mathbb{I}\{L(s_t)> B\}\big)^k|\mF_t]  \\
  & \leq 1 - \eta\alpha + \eta^2 c,
\end{align*}
where we have used the fact that $|\E[\big((L(s_{t+1}) - L(s_{t}))\mathbb{I}\{L(s_t)> B\}\big)^k|\mF_t]| \leq \nu^k$ for all $k \geq 1$, and
notation $c = c(\nu) = e^\nu - \nu - 1$.
Note that since $\eta\in(0, \min\{1, \alpha/(2c)\})$, $1 - \eta\alpha + \eta^2c\leq 1- \eta\alpha/2 = \rho$. To summarize, we obtain that 
for \eqref{eq:lemma:hajek:proof2},
\begin{equation}
    \E[e^{\eta L(s_{t+1})}|\mF_t] \leq e^{\nu + \eta B} +  \rho e^{\eta L(s_t)}. \label{eq:lemma:hajek:proof3}
\end{equation}
By taking expectation on both side of \eqref{eq:lemma:hajek:proof3}, we establish the following recursive equation:
\begin{equation*}
    \E[e^{\eta L(s_{t+1})}] \leq e^{\nu + \eta B} +  \rho \E[e^{\eta L(s_t)}].
\end{equation*}
Since \eqref{eq:lem:hajek:proof1} holds trivially for $t=0$, the above recursive equation implies that the desired inequality 
\eqref{eq:lem:hajek:proof1} is valid for all $t\ge 0$.

Next, we establish \eqref{eq:lem:hajek:recurrent} of Lemma \ref{lem:hajek}. Fix an $a\geq B$ and define $M(t) = \frac{e^{\eta L(s_t)}}{\rho^t}$. Recall that in the above proof, we showed that $\E[e^{\eta (L(s_{t+1}) - L(s_{t}))\mathbb{I}\{L(s_t)> B\}}|\mF_t] \leq \rho$. Therefore, $M(\min(t,T_a))$ is a non-negative supermartingale. This implies that $M(0)\geq \E[M(\min(t,T_a))]$, i.e.,
\begin{align*}
    e^{\eta L(s_0)}&\geq \E[e^{\eta L(s_{\min(t,T_a)}) }/ \rho^{\min(t, T_a)}]\\
    &\geq \E[e^{\eta L(s_{\min(t,T_a)}) }/ \rho^{\min(t, T_a)}\mathbb{I}\{T_a > t\}]\\
    &\geq \frac{e^{\eta a}}{\rho^t}\cdot \P(T_a > t),
\end{align*}
where the last inequality follows from the definition of $T_a$. 
This completes the proof of \eqref{eq:lem:hajek:recurrent} and Lemma \ref{lem:hajek}.

\section{Proof of Theorem \ref{thm:main.2}}
\label{sec:analysis2}

The proof of Theorem \ref{thm:main.2} follows identically as that of Theorem \ref{thm:main.1} by
replacing the performance guarantees of Lemma \ref{lem:sparse} by that of Lemma \ref{lem:sparse.mod}.

\subsection{Proof of Lemma \ref{lem:sparse.mod}}

We establish the statement of Lemma \ref{lem:sparse.mod} inductively. To begin with, we shall
count the total number of samples of the MDP utilized in the algorithm. To that end, note that
for the $H$ steps of the procedure starting with root note $s$, 
any state sampled within $H$ steps from it can not be farther than $H \nu'$ in $\|\cdot\|_2$ 
distance per Condition \ref{cond:lyapunov}. By construction, the number of such states 
contained in $\mS_\varepsilon$ for a given $s$, is at most 
$N(H, \varepsilon) := O((H\nu'/\varepsilon)^d)$. For each of these $N(H, \varepsilon)$ states, 
for each action $a \in \mA$, we need to sample at  most $C$ next states. That is, 
number of samples are at most $C |\mA| N(H, \varepsilon)$. Indeed, as part of our procedure, 
it is likely that some of these $N(H, \varepsilon)$ states are visited multiple times. 

Now, for any $\tilde{s}\in \mS_\varepsilon$  amongst these $N(H, \varepsilon)$ states and for any $a \in \mA$, 
by definition, 
\begin{align}
Q^*(\tilde{s}, a) & = R_{\tilde{s} a} +\gamma \mathbb{E}_{\hat{s} \sim \mathbb{P}(\cdot | \tilde{s}, a)} \big[ V^*(\hat{s})\big].
\end{align}
Let $\{\tilde{s}_i\}_{i \leq C}$ be $C$ sampled next states per MDP at state $\tilde{s}$ under action $a$. Due to the 
standard application of Chernoff's bound for bounded valued variables, for any $\lambda > 0$, 
\begin{align}\label{eq:mc2.1}
\mathbb{P}\Big(\big| \frac{1}{C} \sum_{i=1}^C V^*(\tilde{s}_i) -  \mathbb{E}_{\hat{s} \sim \mathbb{P}(\cdot | \tilde{s}, a)} \big[ V^*(\hat{s})\big] \big| > \lambda\Big)
& \leq 2 \exp\Big(-\frac{\lambda^2 C }{2 V^2_{\max}}\Big). 
\end{align}
Here we have used the fact that $\|V^*\|_\infty \leq V_{\max}$. By choosing 
$\lambda^2 = \frac{2 V^2_{\max}}{C} \log \big(\frac{2 |\mA| N(H, \varepsilon)}{\delta}\big)$, the event within the left-hand side holds with probability at least $1- \frac{\delta}{|\mA| N(H, \varepsilon)}$. Therefore,
by union bound, the event of \eqref{eq:mc2.1} holds for all the $N(H, \varepsilon)$ states and all action $a \in \mA$ pairs. Hence forth
in the remainder of the proof, we shall assume that this event holds. 

Now, for the given query state $s \in \mS$, it is at the root of the sampling tree to produce estimate of 
 $Q^*(s, a)$  for all $a \in \mA$. As part of the procedure, we sample $C$ 
next states for $(s, a)$, which were denoted as $\Sn(s, a, C) = \{s_1,\dots, s_C\}$. We map these states to
their closest elements in $\mS_\varepsilon$, denoted as $S_\varepsilon(s_1),\dots, S_\varepsilon(s_C)$. 
Due to Condition \ref{cond:lip.val}, we have that 
\begin{align}\label{eq:mc2.2}
\big|\frac{1}{C} \sum_{i=1}^C (V^*(s_i) - V^*(S_\varepsilon(s_i)) \big| & \leq \lip \varepsilon \sqrt{d}. 
\end{align}
As per the method, we produce estimate 
\begin{align}
\hQ^0(s, a) & = R_{s a} + \gamma \frac{1}{C} \sum_{i=1}^C \hV^1(S_\varepsilon(s_i)).  
\end{align}
Therefore, we have 
\begin{align}
& \Big| \hQ^0(s, a) - Q^*(s, a) \Big|  \nonumber \\
& \leq \gamma \Big| \frac{1}{C} \sum_{i=1}^C \hV^1(S_\varepsilon(s_i)) - \mathbb{E}_{\tilde{s} \sim \mathbb{P}(\cdot | s, a)} \big[ V^*(\tilde{s})\big] \Big|  \nonumber \\
& \leq \gamma \Big| \frac{1}{C} \sum_{i=1}^C \bigg( \hV^1(S_\varepsilon(s_i)) - V^*(S_\varepsilon(s_i)) + V^*(S_\varepsilon(s_i))  - V^*(s_i) + V^*(s_i) \bigg)  - \mathbb{E}_{\tilde{s} \sim \mathbb{P}(\cdot | s, a)} \big[ V^*(\tilde{s})\big] \Big| \nonumber \\
& \leq \gamma ( \err^1 + \lip \varepsilon \sqrt{d} + \lambda), 
\end{align}
Here $\err^1$ is maximum of error in value function estimates for states in level $1$ in the sampling tree, and
hence $ | \hV^1(S_\varepsilon(s_i)) - V^*(S_\varepsilon(s_i)) | \leq \err^1$ for all $i \leq C$; we have used that
event of \eqref{eq:mc2.1} holds; and \eqref{eq:mc2.2}.  Using this
argument recursively and the fact that $\hV^{h}(\tilde{s}) = \max_{a \in \mA} \hQ^h(\tilde{s}, a)$, 
we have that for all $1\leq h \leq H$,  
\begin{align}
\err^{h-1} & \leq \gamma ( \err^h + \lip \varepsilon \sqrt{d} + \lambda). 
\end{align}
And at the leaf nodes, by definition we have that $\err^H \leq V_{\max}$. Therefore, we conclude that for all $a \in \mA$, 
\begin{align}
\Big| \hQ^0(s, a) - Q^*(s, a) \Big| & \leq \gamma^H V_{\max} + \frac{\gamma ( \lip \varepsilon \sqrt{d} + \lambda)}{1-\gamma}.
\end{align}
We choose $\gamma^H \leq \delta$ or $H = \lceil \log_\gamma(\delta) \rceil$,  
$\varepsilon =  \frac{\delta V_{\max} (1-\gamma)}{2 \lip \gamma \sqrt{d}}$, and $\lambda \leq \frac{\delta V_{\max} (1-\gamma)}{2\gamma}$, i.e. 
\begin{align*}
\frac{2 V^2_{\max}}{C} \log \big(\frac{2 |\mA| N(H, \varepsilon)}{\delta}\big) & \leq \frac{\delta^2 V_{\max}^2 (1-\gamma)^2}{4\gamma^2}, 
\end{align*}
or 
\begin{align*}
C & = \Omega\Big(\frac{\gamma^2}{(1-\gamma)^2 \delta^2} \Big(\log \frac{2 | \mA| }{\delta} + d \log H + d \log \frac{1}{\varepsilon}\Big)\Big)~=~\Omega\Big(\frac{1}{\delta^2} \log \frac{1}{\delta}\Big), 
\end{align*}
where $\Omega(\cdot)$ hides constant dependent on $\gamma, |\mA|, d, V_{\max}$.  Thus, number of
samples utilized are $O(C (H\nu'/\varepsilon)^d) = O\big(\frac{1}{\delta^{2+d}} \log^{d+1} \frac{1}{\delta}\big)$.

\section{Proof of Theorem \ref{thm:main.3}}
\label{sec:analysis3}

Let $G$ be the stopping time defined as $G\triangleq\inf\left\{ t\ge0:\delta_{t}\le\delta(\alpha)\right\} .$
As per the method, we start with $\delta_{0}=1$. If $\delta(\alpha)>1$,
then $G=0$. If $\delta(\alpha)<1,$ then either $G<\infty$ or $G=\infty$.
We consider these two cases separately.

\paragraph{Case 1: $G<\infty$.}

For each $t\ge G$, we have $\delta_{t}\le\delta(\alpha)$. In this
case, the proof of Theorem~\ref{thm:main.1} suggests that the inequality (\ref{eq:thm:stability}) holds;
that is, for all $b\ge0,$
\[
\P(L(s_{t})\geq b\mid s_{G})\leq\rho^{t-G}e^{\eta(L(s_{G})-b)}+\frac{1-\rho^{t-G}}{1-\rho}e^{\nu+\eta(B-b)},
\]
where $c(\nu)=e^{\nu}-\nu-1$, $\eta=\min\left(1,\alpha/4c(\nu)\right)$
and $\rho=1-\eta\alpha/4c(\nu)<1$. By Condition~\ref{cond:lyap.lb}, $L(s)\ge c_{1}\left\Vert s\right\Vert +c_{2},$ hence
\[
\P\left(L(s_{t})\ge b\mid s_{G}\right)\ge\P\left(c_{1}\left\Vert s_{t}\right\Vert +c_{2}\ge b\mid s_{G}\right)=\P\left(\left\Vert s_{t}\right\Vert \ge\frac{b-c_{2}}{c_{1}}\mid s_{G}\right).
\]
Combining the last two display equations and taking $b=c_{1}\log^{2}t+c_{2}$,
we obtain that 
\begin{align*}
\P\left(\left\Vert s_{t}\right\Vert \ge\log^{2}t\mid s_{G}\right) & \le\rho^{t-G}e^{\eta(L(s_{G})-c_{1}\log^{2}t-c_{2})}+\frac{1-\rho^{t-G}}{1-\rho}e^{\nu+\eta(B-c_{1}\log^{2}t-c_{2})}\\
 & \le\underbrace{\left[e^{\eta(L(s_{G})-c_{2})}+\frac{1}{1-\rho}e^{\nu+\eta(B-c_{2})}\right]}_{C\equiv C\left(\nu,\alpha,B,L(s_{G})\right)}e^{-\eta c_{1}\log^{2}t}.
\end{align*}
That is, for each $t\ge G$, the likelihood of test failing at time
$t$ is $Ce^{-\eta c_{1}\log^{2}t}.$ Since $C\sum_{t=G}^{\infty}e^{-\eta c_{1}\log^{2}t}<\infty,$
the Borel-Cantelli lemma ensures that with probability $1$ the test
fails for finitely many times, hence $\lim\inf_{t}\delta_{t}>0$ as
claimed in (\ref{eq:delta_bound}). By definition, when $G<\infty$,
the choice of $\delta$ is such that $\delta\le\delta(\alpha)$. Therefore,
by Theorem 1 or 2, the system is stable.

\paragraph{Case 2: $G=\infty$.}

In this case, the system never reached $\delta\le\delta(\alpha)$
and hence equation (\ref{eq:delta_bound}) holds. In this case, we
may not be able to guarantee stability; nevertheless, we can ensure
near-stability. Now, since we start with $\delta_{0}=1$ and $G=\infty$,
we have $\delta(\alpha)<1$. Since $G=\infty$, the test is failed
no more than $\lceil\log_{2}1/\delta(\alpha)\rceil=O(1)$ times. Therefore,
in the limit of $T\to\infty$, the test is not failed. That is, $\lim\sup_{t\to\infty}\frac{\left\Vert s_{t}\right\Vert }{\log^{2}t}=O(1)$
as claimed in (\ref{eq:limsup_bound}).

\end{document}